\documentclass[11pt,a4paper]{article}
\usepackage[hyperref]{emnlp2018}
\usepackage{times}
\usepackage{latexsym}
\usepackage{algorithm}
\usepackage{algorithmic}
\usepackage{graphicx}
\usepackage{xcolor}
\usepackage{soul}
\usepackage[utf8]{inputenc}
\usepackage[small]{caption}

\usepackage{amsmath}
\usepackage{xcolor}

\usepackage{url}
\usepackage{booktabs}

\usepackage{multirow}
\usepackage{cleveref}
\usepackage{graphicx}
\usepackage{colortbl}
\definecolor{mygray}{gray}{.9}
\Crefformat{section}{\S#2#1#3}

\aclfinalcopy 

\usepackage{xcolor}

\title{SQL-to-Text Generation with Graph-to-Sequence Model}

 \author{Kun Xu$^{1}$\thanks{\quad Work done when the author was at IBM Research.}, Lingfei Wu$^{2}$, Zhiguo Wang$^{2}$,  Yansong Feng$^{3}$, Vadim Sheinin$^{2}$\\
         $^{1}$Tencent AI Lab \\ 
         $^{2}$IBM Research \\
	 $^{3}$Peking University, Beijing, China \\
	\{\texttt{syxu828,zgw.tomorrow}\}\texttt{@gmail.com}, \texttt{lwu@email.wm.edu}\\
	\texttt{fengyansong@pku.edu.cn}, \texttt{vadims@us.ibm.com}
}

\begin{document}

\maketitle
\begin{abstract}
Previous work approaches the SQL-to-text generation task using vanilla Seq2Seq models, which may not fully capture the inherent graph-structured information in SQL query.
In this paper, we first introduce a strategy to represent the SQL query as a directed graph and then employ a graph-to-sequence model to encode the global structure information into node embeddings.
This model can effectively learn the correlation between the SQL query pattern and its interpretation.
Experimental results on the WikiSQL dataset and Stackoverflow dataset show that our model significantly outperforms the Seq2Seq and Tree2Seq baselines, achieving the state-of-the-art performance.
\end{abstract}

\section{Introduction}
\begin{figure}[tb!]
\centering\includegraphics[width=0.45\textwidth]{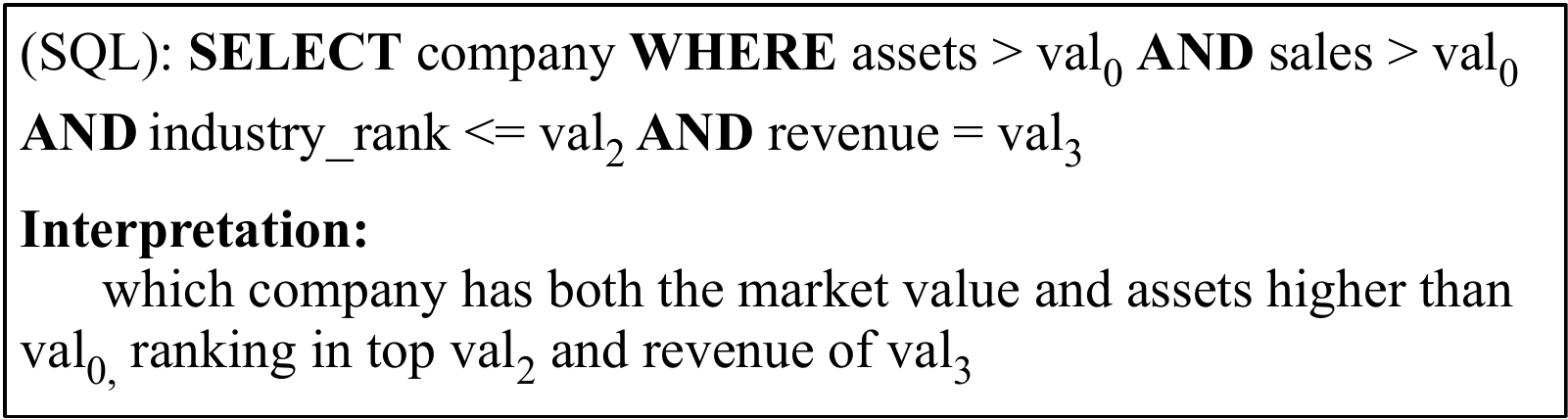}
\caption{An example of SQL query and its interpretation.}
\label{fig:example_intro}
\end{figure}

The goal of the SQL-to-text task is to automatically generate human-like descriptions interpreting the meaning of a given structured query language (SQL) query (Figure~\ref{fig:example_intro} gives an example).
This task is critical to the natural language interface to a database since it helps non-expert users to understand the esoteric SQL queries that are used to retrieve the answers through the question-answering process \cite{simitsis2009dbmss} using varous text embeddings techniques \cite{kim2014convolutional,arora2017simple,wordwu2018}.

Earlier attempts for SQL-to-text task are rule-based and template-based \cite{koutrika2010explaining,ngonga2013sorry}.
Despite requiring intensive human efforts to design temples or rules, these approaches still tend to generate rigid and stylized language that lacks the natural text of the human language.
To address this, \newcite{iyer2016summarizing} proposes a sequence-to-sequence (Seq2Seq) network to model the SQL query and natural language jointly.
However, since the SQL is designed to express graph-structured query intent,
the sequence encoder may need an elaborate design to fully capture the global structure information.
Intuitively, varous graph encoding techniques base on deep neural network \cite{kipf2016semi,DBLP:conf/nips/HamiltonYL17,song2018graph} or based on Graph Kernels \cite{vishwanathan2010graph,wu2018d2ke}, whose goal is to learn the node-level or graph-level representations for a given graph, are more proper to tackle this problem.

In this paper, we first introduce a strategy to represent the SQL query as a directed graph (see \Cref{sec:SQL Query Transformation}) and further make full use of a novel graph-to-sequence (Graph2Seq) model \cite{xu2018graph2seq}
that encodes this graph-structured SQL query, and then decodes its interpretation (see \Cref{sec:graph2seq}).
On the encoder side, we extend the graph encoding work of \newcite{DBLP:conf/nips/HamiltonYL17} by encoding the edge direction information into the node embedding.
Our encoder learns the representation of each node by aggregating information from its $K$-hop neighbors.
Different from \newcite{DBLP:conf/nips/HamiltonYL17} which neglects the edge direction, we classify the neighbors of a node according to the edge direction, say $v$, into two classes, i.e., forward nodes ($v$ directs to) and backward nodes (direct to $v$).
We apply two distinct aggregators to aggregate the information of these two types of nodes, resulting two representations.
The node embedding of $v$ is the concatenation of these two representations.
Given the learned node embeddings, we further introduce a pooling-based and an aggregation-based method to generate the graph embedding.

On the decoder side, we develop an RNN-based decoder which takes the graph vector representation as the initial hidden state to generate the sequences while employing an attention mechanism over all node embeddings.
Experimental results show that our model achieves the state-of-the-art performance on the WikiSQL dataset and Stackoverflow dataset.
Our code and data is available at {\small \url{https://github.com/IBM/SQL-to-Text}}.

\section{Graph Representation of SQL Query}
\label{sec:SQL Query Transformation}
Representing the SQL query as a graph instead of a sequence could better preserve the inherent structure information in the query.
An example is illustrated in the blue dashed frame in Figure 2.
One can see that representing them as a graph instead of a sequence could help the model to better learn the correlation between this graph pattern and the interpretation ``...\textit{both X and Y higher than Z}...''.
This observation motivates us to represent the SQL query as a graph.
In particular, we use the following method to transform the SQL query to a graph:\footnote{This method could be simply extended to cope with more general SQL queries that have complex syntaxes such as JOIN and ORDER BY.}

\textbf{SELECT Clause.}
For the SELECT clause such as ``SELECT company'', we first create a node assigned with text attribute \textit{select}. This SELECT node connects with column nodes whose text attributes are the selected column names such as \textit{company}.
For SQL queries that contain aggregation functions such as \texttt{count} or \texttt{max}, we add one aggregation node which is connected with column nodes.
Similarly, their text attributes are the aggregation function names.

\begin{figure}[t!]
\centering\includegraphics[width=0.38\textwidth]{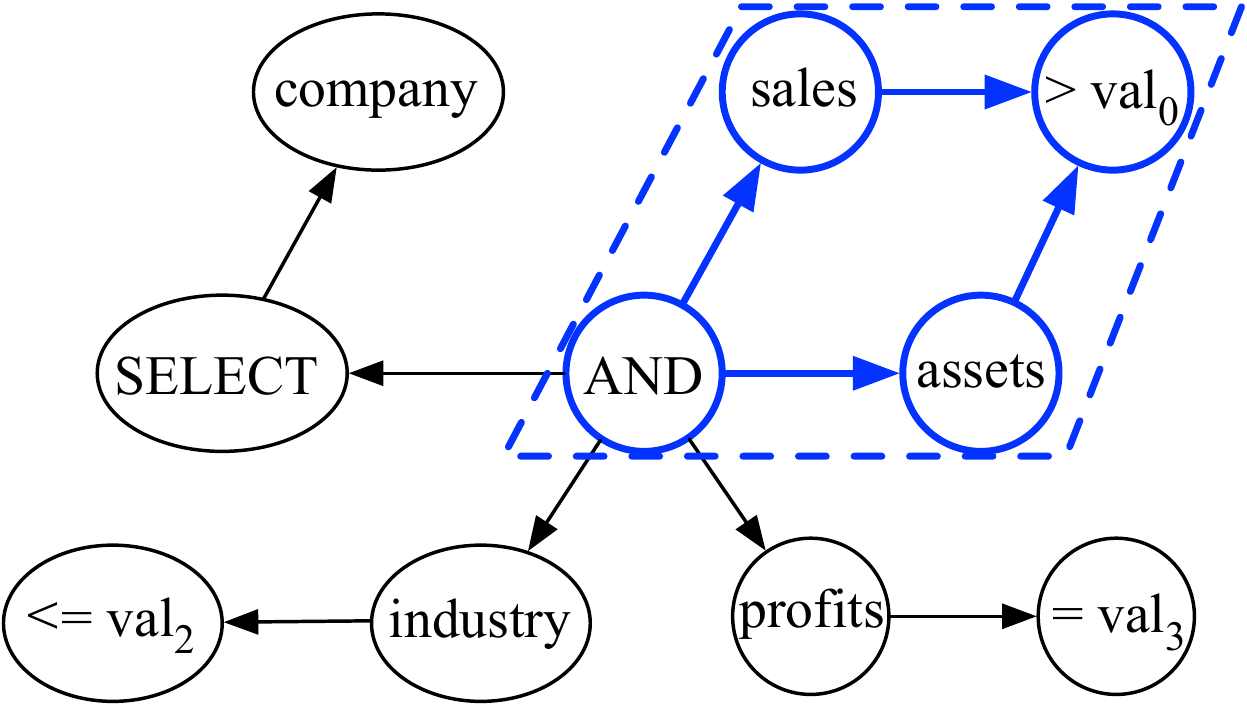}
\caption{The graph representation of the SQL query in Figure~\ref{fig:example_intro}.}
\label{fig:example}
\end{figure}

\textbf{WHERE Clause.}
The WHERE clause usually contains more than one condition.
For each condition, we use the same process as for the SELECT clause to create nodes.
For example, in Figure~\ref{fig:example}, we create node \textit{assets} and $>$\textit{$val_0$} for the first condition,
the node \textit{sales} and $>$\textit{$val_0$} for the second condition.
We then integrate the constraint nodes that have the same text attribute (e.g., $>$\textit{$val_0$} in Figure~\ref{fig:example}).
For a logical operator such as AND, OR and NOT,
we create a node that connects with all column nodes that the operator works on.
Finally, these logical operator nodes connect with the SELECT node.

\section{Graph-to-sequence Model}
\label{sec:graph2seq}
Based on the constructed graphs for the SQL queries, we make full use of a novel graph-to-sequence model \cite{xu2018graph2seq}, which consists of a graph encoder to learn the embedding for the graph-structured SQL query, and a sequence decoder with attention mechanism to generate sentences.
Conceptually, the graph encoder generates the node embedding for each node by accumulating information from its $K$-hop neighbors, and produces a graph embedding for the entire graph by abstracting all node embeddings. Our decoder takes the graph embedding as the initial hidden state and calculates the attention over all node embeddings on the encoder side to generate natural language interpretations. 

\paragraph{Node Embedding.}
Given the graph $\mathcal{G} = (\mathcal{V}, \mathcal{E})$,
since the text attribute of a node may include a list of words, 
we first use a Long Short Term Memory (LSTM) to generate the feature vector \textbf{a}$_{v}$ for all nodes  $\forall v \in \mathcal{V}$ from $v$'s text attribute.
We use these feature vectors as initial node embeddings.
Then, our model incorporates information from a node's neighbors within $K$ hop into its representation by repeating the following process $K$ times:
\begin{gather}
\small
\textbf{h}_{v\vdash}^{0} = \textbf{a}_{v}, \textbf{h}_{\vdash v}^{0} = \textbf{a}_{v}, \forall v \in \mathcal{V} \\
\textbf{h}_{\mathcal{N}_{\vdash}(v)}^{k}   = \textbf{M}_{\vdash}^{k}(\{\small \textbf{h}_{u\vdash}^{k-1}, \forall u \in \mathcal{N}_{\vdash}(v)\}) \\
\textbf{h}_{v\vdash}^{k} = \sigma ( \textbf{W}^{k}\cdot \texttt{\small CONCAT}(\textbf{h}_{v\vdash}^{k-1}, \textbf{h}_{\mathcal{N}_{\vdash}(v)}^{k})) \\
\textbf{h}_{\mathcal{N}_{\dashv}(v)}^{k}   = \textbf{M}_{\dashv}^{k}(\{\small \textbf{h}_{u\dashv}^{k-1}, \forall u \in \mathcal{N}_{\dashv}(v)\}) \\
\textbf{h}_{v\dashv}^{k} = \sigma ( \textbf{W}^{k}\cdot \texttt{\small CONCAT}(\textbf{h}_{v\dashv}^{k-1}, \textbf{h}_{\mathcal{N}_{\dashv}(v)}^{k}))
\end{gather}
where $k\in\{1,...,K\}$ is the iteration index,
$\mathcal{N}$ is the neighborhood function\footnote{$\mathcal{N}_{\vdash}(v)$ returns the nodes that $v$ directs to and $\mathcal{N}_{\dashv}(v)$ returns the nodes that direct to $v$.}, 
\textbf{h}$_{v\vdash}^{k}$ (\textbf{h}$_{v\dashv}^{k}$) is node $v$'s forward (backward) representation 
which aggregates the information of nodes in $\mathcal{N}_{\vdash}(v)$ ($\mathcal{N}_{\dashv}(v)$),
\textbf{M}$_{\vdash}^{k}$ and \textbf{M}$_{\dashv}^{k}$ are the forward and backward aggregator functions, \textbf{W}$^{k}$ denotes weight matrices, $\sigma$ is a non-linearity function.

For example, for node $v\in \mathcal{V}$, we first aggregate the forward representations of its immediate neighbors \{\textbf{h}$_{u\vdash}^{k-1}$, $\forall u \in \mathcal{N}_{\vdash}(v)$\} into a single vector \textbf{h}$_{\mathcal{N}_{\vdash}(v)}^{k}$ (equation 2).
Note that this aggregation step only uses the representations generated at previous iteration and its initial representation is \textbf{a}$_{v}$.
Then we concatenate $v$'s current forward representation \textbf{h}$_{v\vdash}^{k-1}$ with the newly generated neighborhood vector \textbf{h}$_{\mathcal{N}_{\vdash}(v)}^{k}$. This concatenated vector is fed into a fully connected layer with nonlinear activation function $\sigma$,
which updates the forward representation of $v$ to be used at the next iteration (equation 3).
Next, we update the backward representation of $v$ in the similar fashion (equation 4$\sim$5).
Finally, the concatenation of the forward and backward representation at last iteration $K$, is used as the resulting representation of $v$.
Since the neighbor information from different hops may have the different impact on the node embedding,
we learn a distinct aggregator function at each step.
This aggregator feeds each neighbor's vector to a fully-connected neural network and an element-wise
max-pooling operation is applied to capture different aspects of the neighbor set.

\paragraph{Graph Embedding.}
Most existing works of graph convolution neural networks focus more on node embeddings rather than graph embeddings (GE) since their focus is on the node-wise classification task.
However, graph embeddings that convey the entire graph information are essential to the downstream decoder, which is crucial to our task.
For this purpose, we propose two ways to generate graph embeddings, namely, the Pooling-based and Node-based methods.

\textbf{Pooling-based GE.}
This method feeds the obtained node embeddings into a fully-connected neural network and applies the element-wise
\textit{max}-pooling operation on all node embeddings.
In experiments, we did not observe significant performance improvement using min-pooling and average-pooling.

\textbf{Node-based GE.}
Following \cite{scarselli2009graph}, this method adds a \textbf{super} node $v_{s}$ that is connected to all other nodes by a special type of edge.
The embedding of $v_{s}$, which is treated as graph embedding, is produced using node embedding generation algorithm mentioned above.

\paragraph{Sequence Decoding.}
The decoder is an RNN which predicts the next token $y_{i}$ given all the previous words
$y_{<i} = y_{1},...,y_{i-1}$, the RNN hidden state $s_{i}$ for time-step $i$ and
the context vector $c_{i}$ that captures the attention of the encoder side.
In particular, the context vector $c_{i}$ depends on a set of node representations (\textbf{h$_{1}$},...,\textbf{h$_{\mathcal{V}}$})
to which the encoder maps the input graph.
The context vector $c_{i}$ is dynamically computed using attention mechanism over the node representations.
Our model is jointly trained to maximize the conditional log-probability of the correct description given 
a source graph with respect to the parameters $\theta$ of the model:
\begin{displaymath}
\small
\theta^{*} = \arg \max_{\theta}\sum_{n=1}^{N}\sum_{t=1}^{T_{n}}\log p(y_{t}^{n}|y_{<t}^{n},x^{n})
\end{displaymath}
where ($x^{n}$, $y^{n}$) is the $n$-th SQL-interpretation pair in the training set, and $T_{n}$ is the length of the $n$-th target sentence $y^{n}$.
In the inference phase, we use the beam search algorithm with beam size = 5.
\section{Experiments}
We evaluate our model on two datasets, WikiSQL \cite{zhongSeq2SQL2017} and Stackoverflow \cite{iyer2016summarizing}.
WikiSQL consists of a corpus of 87,726 hand-annotated SQL query and 
natural language question pairs.
These SQL queries are further split into training (61,297 examples), development (9,145 examples) and test sets (17,284 examples).
StackOverflow consists of 32,337 SQL query and natural language question pairs, and we use the same train/development/test split as \cite{iyer2016summarizing}.
We use the BLEU-4 score \cite{DBLP:conf/acl/PapineniRWZ02} as our automatic evaluation metric and also perform a human study.
For human evaluation, we randomly sampled 1,000 predicted results and
asked three native English speakers to rate each interpretation against both the correctness conforming to the input SQL and grammaticality on a scale between 1 and 5.
We compare some variants of our model against the template, Seq2Seq, and Tree2Seq baselines.

\textbf{\textit{Graph2Seq}-PGE.}
This method uses the \textbf{P}ooling method for generating \textbf{G}raph \textbf{E}mbedding.

\textbf{\textit{Graph2Seq}-NGE.}
This method uses the \textbf{N}ode based \textbf{G}raph \textbf{E}mbedding.

\textbf{Template.} We implement a template-based method which first maps each element of a SQL query to an utterance and then uses simple rules to assemble these utterances. For example, we map SELECT to \textit{which}, WHERE to \textit{where}, $>$ to \textit{more than}.
This method translates the SQL query of Figure~\ref{fig:example_intro} to \textit{which company where assets more than val$_0$ and sales more than val$_0$ and industry less than or equal to val$_1$ and profits equals val$_{2}$}.

\textbf{Seq2Seq.}
We choose two Seq2Seq models as our baselines.
The first one is the attention-based Seq2Seq model proposed by \newcite{DBLP:journals/corr/BahdanauCB14},
and the second one additionally introduces the copy mechanism in the decoder side \citep{gu2016incorporating}.
To evaluate these models, we employ a template to convert the SQL query into a sequence:
``{\small SELECT + \textit{$<$aggregation function$>$} + $<$Split Symbol$>$ + \textit{$<$selected column$>$} + WHERE + \textit{$<$condition$_{0}$$>$} + $<$Split Symbol$>$ + \textit{$<$condition$_{1}$$>$ + ...}} ''.

\textbf{Tree2Seq.}
We also choose a tree-to-sequence model proposed by \cite{eriguchi2016tree} as our baseline.
We use the SQL Parser tool\footnote{\url{http://www.sqlparser.com}} to convert a SQL query into the tree structure\footnote{See Appendix for details.} which is fed to the Tree2Seq model.

Our proposed models are trained using the Adam optimizer \cite{DBLP:journals/corr/KingmaB14},
with mini-batch size 30.
Our hyper-parameters are set based on performance on the validation set.
The learning rate is set to 0.001.
We apply the dropout strategy \cite{DBLP:journals/jmlr/SrivastavaHKSS14} with the ratio of 0.5 at the decoder layer to avoid overfitting.
Gradients are clipped when their norm is bigger than 20.
We initialize word embeddings using GloVe word vectors from \newcite{pennington2014glove}, and the 
word embedding dimension is 300.
For the graph encoder, the hop size $K$ is set to 6, the non-linearity function $\sigma$ is implemented as ReLU \cite{DBLP:journals/jmlr/GlorotBB11},
the parameters of weight matrices \textbf{W}$^{k}$ are randomly initialized.
The decoder has one layer, and its hidden state size is 300.

\paragraph{Results and Discussion}
\begin{table}[t!]
\small
\centering
\begin{tabular}{cccc}
 & BLEU-4 & Grammar. & Correct.  \\
\toprule[0.8pt]
Template & 15.71 & 1.50 &  - \\
Seq2Seq & 20.91 &  2.54 & 62.1\% \\
Seq2Seq + Copy & 24.12 & 2.65 & 64.5\% \\
Tree2Seq   & 26.67 & 2.70 & 66.8\% \\
\rowcolor{mygray}
\textit{Graph2Seq}-PGE &  \textbf{38.97} & \textbf{3.81} & \textbf{79.2\%} \\
\rowcolor{mygray}
\textit{Graph2Seq}-NGE & 34.28 & 3.26 & 75.3\% \\
\hline
\hline
\cite{iyer2016summarizing} & 18.4 & 3.16 & 64.2\% \\
\rowcolor{mygray}
\textit{Graph2Seq}-PGE & \textbf{23.3} & \textbf{3.23} & \textbf{70.2\%} \\
\rowcolor{mygray}
\textit{Graph2Seq}-NGE & 21.9 & 2.97 & 65.1\%\\
\toprule[0.8pt]
\end{tabular}
\caption{Results on the WikiSQL (above) and Stackoverflow (below).}
\label{tab:results}
\end{table}
Table~\ref{tab:results} summarizes the results of our models and baselines.
Although the template-based method achieves decent BLEU scores, its grammaticality score is substantially worse than other baselines.
We can see that on both two datasets, our Graph2Seq models perform significantly better than the Seq2Seq and Tree2Seq baselines.
One possible reason is that in our graph encoder, the node embedding retains the information of neighbor nodes within $K$ hops. However, in the tree encoder, the node embedding only aggregates the information of descendants while losing the knowledge of ancestors.
The pooling-based graph embedding is found to be more useful than the node-based graph embedding because \textit{Graph2Seq}-NGE adds a nonexistent node into the graph, which introduces the noisy information in calculating the embeddings of other nodes. We also conducted an experiment that treats the SQL query graph as an undirected graph and found the performance degrades.

By manually analyzing the cases in which the Graph2Seq model performs better than Seq2Seq,
we find the Graph2Seq model is better at interpreting two classes of queries: (1) the complicated queries that have more than two conditions (Query 1); (2) the queries whose columns have implicit relationships (Query 2).
Table~\ref{tab:examples} lists some such SQL queries and their interpretations.
One possible reason is that the Graph2Seq model can better learn the correlation between the graph pattern and natural language by utilizing the global structure information.

\begin{table}[t]
\small
\centering
\begin{tabular}{l}
\toprule[0.8pt]
\multicolumn{1}{c}{SQL Query \& Interpretations} \\
\hline
\rowcolor{mygray} $1$. COUNT Player WHERE starter = val$_{0}$ AND touchdowns\\
\rowcolor{mygray} \quad  = val$_{1}$ AND position = val$_{2}$ \\
\textbf{S}: How many players played in position val$_{2}$ \\
\textbf{G}: number of players with starter val$_{0}$ and get touchdowns \\ 
val$_{1}$ for val$_{2}$
\\ 
\rowcolor{mygray} $2$. SELECT Tires WHERE engine = val$_0$ AND chassis = \\
\rowcolor{mygray} \quad  val$_1$ AND team $=$ val$_2$ \\
\textbf{S}: which tire has engine val$_0$ and chassis val$_1$ and val$_{2}$ \\ 
\textbf{G}: which tire does val$_2$ run with val$_{0}$ engine and val$_1$ chassis\\
\toprule[0.8pt]
\end{tabular}
\caption{Example of SQL queries and predicted interpretations where S and G denotes Seq2Seq and Graph2Seq models, respectively.}
\label{tab:examples}
\end{table}

We find the hop size has a significant impact on our model since it determines
how many neighbor nodes to be considered during the node embedding generation.
As the hop size increasing, the performance is found to be significantly improved.
However, after the hop size reaches 6, increasing the hop size can not boost the performance on WikiSQL anymore.
By analyzing the most complicated queries (around 6.2\%) in WikiSQL, we find there are average six hops between a node and its most distant neighbor.
This result indicates that the selected hop size should guarantee each node can receive the information from others nodes in the graph.

\section{Conclusions}
Previous work approaches the SQL-to-text task using an Seq2Seq model which does not fully capture the global structure information of the SQL query.
To address this, we proposed a Graph2Seq model which includes a graph encoder, an attention based sequence decoder.
Experimental results show that our model significantly outperforms the Seq2Seq and Tree2Seq models on the WikiSQL and Stackoverflow datasets.

\section*{Appendix}
\begin{figure}[h]
\centering\includegraphics[width=0.45\textwidth]{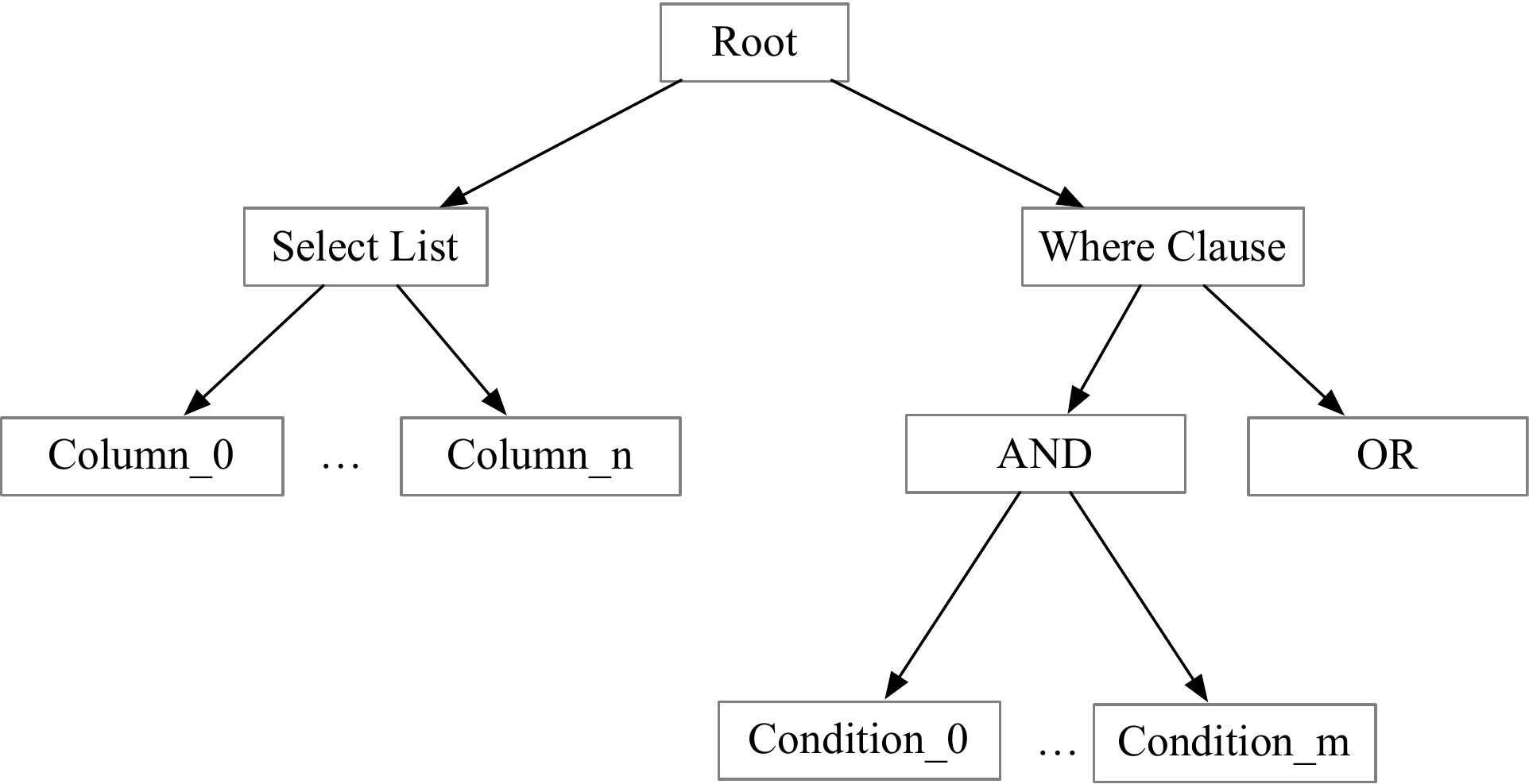}
\caption{Tree representation of the SQL query.}
\label{fig:tree_rep}
\end{figure}
We apply the SQL Parser tool\footnote{\url{http://www.sqlparser.com}} to convert an SQL query to a tree whose structure is illustrated in Figure~\ref{fig:tree_rep}. More specifically, the root has two child nodes, namely \underline{\textit{Select List}} and \underline{\textit{Where Clause}}. The child nodes of \underline{\textit{Select List}} represent the selected columns in the SQL query. The \underline{\textit{Where Clause}} has the logical operators occurred in the SQL query as its children. The children of a logical operator node are the conditions on which this operator works. 

\bibliographystyle{acl_natbib}
\bibliography{emnlp2018}

\newpage
\appendix

\end{document}